\documentclass[conference]{IEEEtran}
\IEEEoverridecommandlockouts

\usepackage{cite}
\usepackage{amsmath,amssymb,amsfonts}
\usepackage{mathrsfs}
\usepackage{algorithm2e}
\usepackage{algpseudocode}
\usepackage{textcomp}
\usepackage[utf8]{inputenc}
\usepackage[english]{babel}
\usepackage[pdftex]{graphicx}
\usepackage{pgfplots}
\pgfplotsset{compat=1.17}
\usepackage{pgfplotstable}
\usepackage{filecontents}
\usepackage{subcaption}
\usepackage{eso-pic}

\newcommand{\lSec}[1]{\label{sec:#1}}
\newcommand{\rSec}[1]{Section \ref{sec:#1}}

\usepackage{xcolor}
\def\BibTeX{{\rm B\kern-.05em{\sc i\kern-.025em b}\kern-.08em
    T\kern-.1667em\lower.7ex\hbox{E}\kern-.125emX}}
\begin{document}


\newcommand\AtPageUpperMycenter[1]{\AtPageUpperLeft{%
 \put(\LenToUnit{0.05\paperwidth},\LenToUnit{-1cm}){%
     \parbox{0.8\textwidth}{\raggedleft\fontsize{9}{11}\selectfont #1}}%
 }}%
\newcommand{\conf}[1]{%
\AddToShipoutPictureBG*{%
\AtPageUpperMycenter{#1}
}
}

\newcommand{\system}{$F^3S$}

\title{\system: Free Flow Fever Screening\\
{\footnotesize \textsuperscript{}}
\thanks{\textsuperscript{*} Work done while being affiliated with NEC Laboratories America, Inc.\\2693-8340/21/\$31.00 $\copyright$2021 IEEE\\DOI 10.1109/SMARTCOMP52413.2021.00060}
}


\author{\IEEEauthorblockN{Kunal Rao, Giuseppe Coviello, Min Feng\textsuperscript{*}, Biplob Debnath, Wang-Pin Hsiung, \\ Murugan Sankaradas, Yi Yang\textsuperscript{*}, Oliver Po, Utsav Drolia\textsuperscript{*} and Srimat Chakradhar}
\IEEEauthorblockA{
\textit{NEC Laboratories America, Inc.}\\
Princeton, NJ and San Jose, CA}
}

\maketitle
\conf{2021 IEEE International Conference on Smart Computing (SMARTCOMP)}

\begin{abstract}
Identification of people with elevated body temperature can reduce or dramatically slow down the spread of infectious diseases like COVID-19. We present a novel fever-screening system, \system, that uses edge machine learning techniques to accurately measure core body temperatures of multiple individuals in a free-flow setting. \system\ performs real-time sensor fusion of visual camera with thermal camera data streams to detect elevated body temperature, and it has several unique features: (a) visual and thermal streams represent very different modalities, and we dynamically associate semantically-equivalent regions across visual and thermal frames by using a new, dynamic alignment technique that analyzes content and context in real-time, (b) we track people through occlusions, identify the eye (inner canthus), forehead, face and head regions where possible, and provide an accurate temperature reading by using a prioritized refinement algorithm, and (c) we robustly detect elevated body temperature even in the presence of personal protective equipment like masks, or sunglasses or hats, all of which can be affected by hot weather and lead to spurious temperature readings. \system\ has been deployed at over a dozen large commercial establishments, providing contact-less, free-flow, real-time fever screening for thousands of employees and customers in indoors and outdoor settings.
\end{abstract}


\section{Introduction}
\begin{figure}[ht]
\centering
\includegraphics[width=0.6\linewidth]{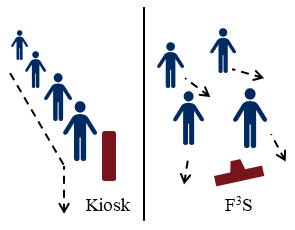}
\caption{Pause-and-Go (left) vs. Free-Flow (right)}
\label{fig:pause-vs-free-flow}
\end{figure}

\lSec{introduction}
One of the common symptoms for a person infected with COVID-19 is fever~\cite{cdc}. Reliable and accurate detection of fever helps in isolating potentially infected people. In this paper, we present \system\, which screens for people with fever as they move in a free flow setting where individuals need not pause or stop at a kiosk for fever screening. Fig. \ref{fig:pause-vs-free-flow} on the left shows a setup where people need to pause or stop, get their temperature measured and then proceed, while on the right, it shows the free flow movement of people, where temperature is measured as people walk through. By allowing free flow movement, \system\ increases the throughput (number of people screened per minute) and avoids over-crowding (which can increase virus transmission).

To measure temperature of individuals, \system\ uses thermal and visual cameras, which are placed far away from people, so that temperature measurements can be done from a distance in a contact-less manner. This does not require any human intervention, which is a key concern for COVID-19, since it can spread through close contact of individuals \cite{cdc}. To do simultaneous fever screening of multiple persons in a free-flow setting, and in real-time, \system\ uses edge machine learning techniques to perform sensor fusion of thermal  and visual data streams within a few hundred milliseconds. Such low-latency response for multiple persons is not possible when visual and thermal data streams are sent to the cloud for analytics processing. \system\ uses resources near the visual and thermal sensors for edge machine learning.

We make the following key contributions:
\begin{enumerate}
  \item We present \system, a novel free flow fever screening solution which runs at the edge and through use of deep learning techniques, it enables real-time, simultaneous high-throughput measurement of core body temperature of multiple individuals from a distance, all without any human intervention.
  \item We present novel sensor fusion techniques to fuse thermal and visual frames to enable accurate temperature measurement of multiple individuals simultaneously.
  \item We present a neural network based distance compensation model, which enables correct temperature measurement at various distances from the thermal camera. Such compensation is necessary because the temperature reported for a person depends on the distance of the person from the thermal camera. 
  \item We present novel techniques to track temperature of individuals across frames, and at different regions (eye, forehead, face and head), depending on visibility and pose of the person; we prioritize, collate and filter alerts for same individual to avoid false positives.
  \item We present novel techniques to measure temperature of individuals even when their face is partially covered (for example, if they are wearing  masks, sunglasses or hats)
  \item Finally, we present a methodology to determine ground truth using thermal and visual sensor data, and to verify accuracy and correctness of temperature measurement.
\end{enumerate}

\section{Background and Challenges}
Fever screening measures the core body temperature of individuals as they walk into the facility and triggers an alert when the temperature is above a certain pre-configured threshold (Centers for Disease Control and Prevention considers a  person to have fever when measured temperature is at least 100.4$^{\circ}$F or 38$^{\circ}$C~\cite{cdc}). 

The temperature across a human body is not consistent -- head is the hottest, feet are the coldest, and there are variations in between.  
There are two regions of the human body that can provide most reliable core temperature measurements: 1) inner canthus of the eye -- corner of the eye where the upper and lower eyelids meet, and 2) ear canals. For non-invasive febrile temperature screening using thermographic devices, ISO recommends obtaining temperature reading between the eyes of a person~\cite{ISO/TR13154:2017}.

This is the main goal associated with measuring core body temperatures remotely, i.e. non-invasively and without human proximity, using thermographic sensors - detecting this region reliably, and measuring accurate core body temperature.

There are two main challenges that \system\ addresses in achieving this goal.
\begin{enumerate}
 \item \textbf{Free Flow:} How to reliably measure core body temperature while people are moving, possibly occluding each other, and are not asked to stop and look at the sensor?
 
 \item \textbf{Eye-region Occlusion:} How to adapt to eye-region obstructions, such as glasses, caps, masks; measure core body temperature and still allow uninterrupted flow of entry?

 
 
\end{enumerate}

\section{Related Work}
Infrared thermography has been deployed at the quarantine stations for detecting elevated body temperature in the last two decades~\cite{contactless-vital-signs-2020, applications-irt-2017, infrared-thermal-detection-comparisons-2015}. There are many research works that combine RGB and thermal images to extract multiple vital signs (i.e. body temperature, heart rate, respiration rate, blood volume pulse, etc.) to identify people with febrile conditions~\cite{contactless-vital-signs-2020, stable-contactles-sensing-with-face-2018, remote-sensing-CMOS-2017, multiple-vital-sign-comparisons-2016}. However, these works have been designed for the pause-and-go like scenario, where an individual pauses for a few seconds, stands still and looks straight at the camera. Even commercial solutions like \cite{ici-temperature-screening, lamasatech, ignatiuz, kriosk, fisherstech, diversified} work well only in pause-and-go scenarios and not in a free-flow setting. They assume that the face and eye-region is clearly visible with no obstruction, which may not be the case in a free-flow scenario, where people are screened as they walk through and face and/or eye-region may or may not be always visible due to occlusion.

Haghmohammadi et al. propose a method for measuring body temperature for an indoor moving crowd~\cite{body-tempature-crowd-2018}. However, it also does not work when the face is not clearly visible. In addition, it has not been tested thoroughly for crowd scenario. Somboonkaew et al. propose a mobile-platform for massive human temperature screening in large public areas based on the infrared forehead temperature using an IR camera and a mobile phone~\cite{mobile-platform-2017}. They align the RGB image to the thermal image to find the target area of the temperature measurement. For the image alignment, they use image cropping and image scaling based on field of view and image resolution of the RGB and thermal cameras. This alignment procedure, however, is very simple and does not work well when faces are occluded. In contrast, \system\ uses sophisticated method to dynamically align visual and thermal image pairs (explained in \rSec{dynamic-frame-alignment}), which enables accurate temperature measurement in a free-flow setting, even when there is occlusion.
\section{\system\ Overview}
\system\ is a fever screening system designed to operate in a free-flow manner i.e. individuals are not required to pause/stop in order for the temperature to be measured. Instead, their temperatures is measured automatically from a distance as they cross the field of view of the camera. A key aspect of our system is that it does not require any human intervention, which is a key concern for COVID-19, since it can spread through close contact of individuals \cite{cdc}. Our solution works in real-time and is able to handle a high volume and rate of flow.

\begin{figure}[b]
\centering
\includegraphics[width=0.9\linewidth]{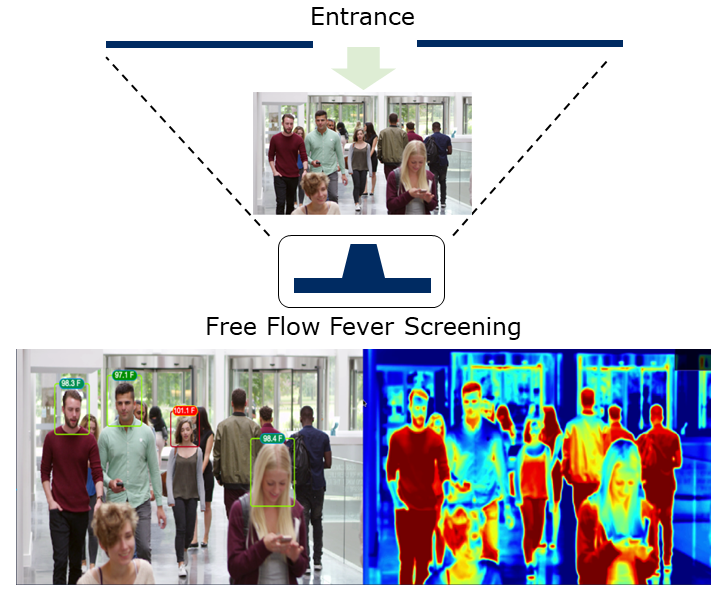}
\caption{\system\ Deployment}
\label{fig:deployment}
\end{figure}

Fig. \ref{fig:deployment} shows the setup for deployment of \system. Arrow shows the direction of movement of people from the entrance into the building. Cameras are located further away pointing towards the entrance so that a large enough field-of-view is captured. As people walk into the building, their temperatures are measured and displayed on the screen for the operator to monitor.

If the temperature of a person is above the configured threshold, then an alert is triggered and the operator can request the individual to step aside and proceed for secondary screening. \system\ acts as an initial screening solution and final screening is done using a medical-grade thermometer. Physical distance between operator and target individuals is constantly maintained, as the operator does not need to come in close contact with the people. In fact, the operator need not be present at the location physically as well, everything can be monitored from a central location in the control room and if an alert is triggered, the individual can be notified over an audio speaker to step aside, thereby avoiding the need for any human to physically intervene.


\section{\system\ Design}
\lSec{design}
The primary goal of any fever screening system is to measure a person's temperature in their eye region, particularly inner canthus area, to produce the most reliable measurement \cite{ISO/TR13154:2017}. Using a thermal sensor alone to detect this region and measure the temperature is possible, but the accuracy of detection is poor \cite{jangblad2018object}.
In order to overcome this limitation, we combine visual sensor data with thermal sensor data to get accurate readings of the temperature of every person. 
Visual sensors produce higher dimensional data than thermal sensors (higher resolution and channels in image data). Therefore, visual camera streams allow for more accurate detection of persons, their faces and eye-regions. However, visual cameras lack information about temperature, which is available in the thermal stream data. We use deep learning models to perform sensor fusion of visual and thermal streams to associate temperatures in thermal data with corresponding objects in the visual stream.

\begin{figure}[b]
\centering
\includegraphics[width=0.99\linewidth]{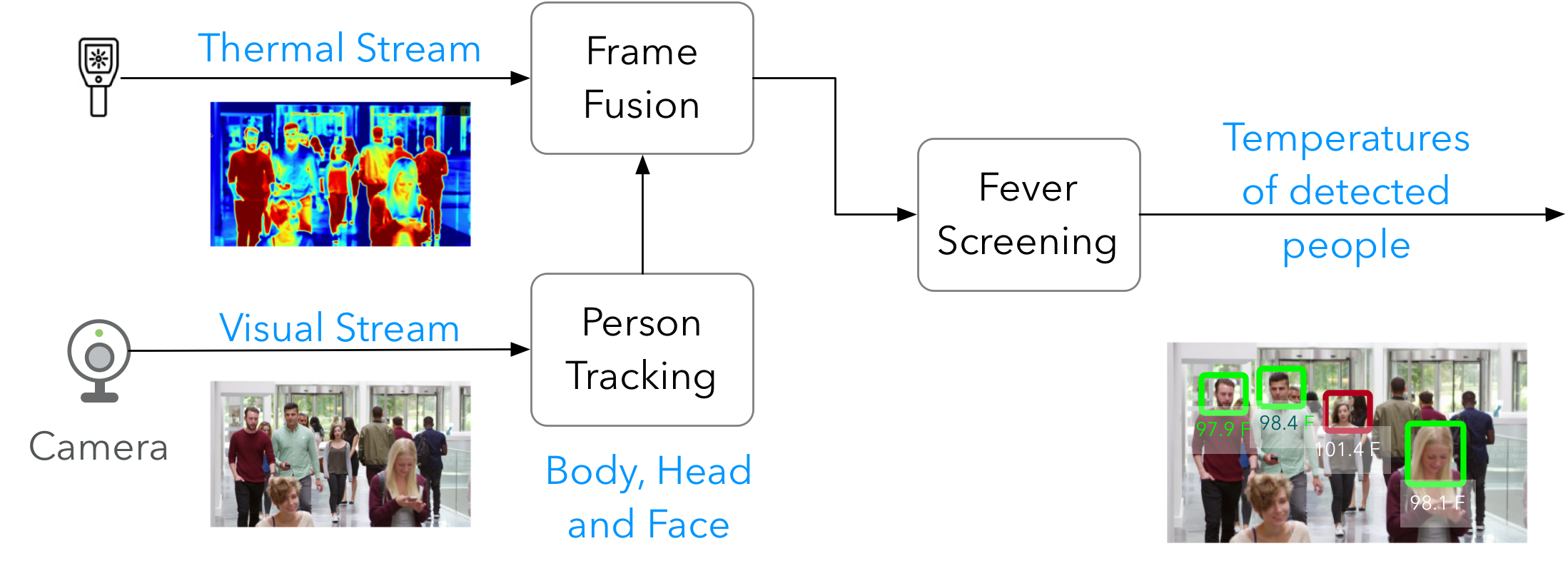}
\caption{\system\ design and workflow}
\label{fig:system-design}
\end{figure}

Fig. \ref{fig:system-design} shows a high-level design and workflow of \system. There are two streams of data coming into our system i.e. thermal data stream and visual data stream. Both these streams relate to the same scene, one containing the visual RGB frame and the other containing the thermal frame of the scene. 

\noindent\textbf{Goal.} The system's goal is to detect the core body temperature of people as they move in a free-flow manner, from their eye region when possible, or other appropriate body regions.




In order to achieve this goal, we split the design of \system\ into three components, each of them handling specific challenges. 
\begin{enumerate}
    \item \textbf{Person Tracking:} The system needs to detect and track people, while also identifying various regions for temperature measurement as they walk through.
    \item \textbf{Frame Fusion:} The system needs to correlate the identified regions in the visual frame with regions in the thermal frame, from where temperatures can be extracted.
    \item \textbf{Fever Screening:} The system needs to decide if a person actually has fever or not, based on the multiple detections it possibly makes across frames.
\end{enumerate} 


Next, we discuss each of these components in detail.

\subsection{Person Tracking}
This component of \system\ detects people and tracks them as they move in a free-flow setting. We use a proprietary neural network based model \cite{neoface}, which detects the face, head and body of a person. Depending on whether the person is wearing glasses, mask, hat, or a combination of these, and the pose and angle - none, one or more of the above detections are possible for a person in each frame. After detection, tracking of the person is a key component in our system to achieve high accuracy when people are not pausing or stopping. Having a unique tracking ID for the detected regions of a person allows our system to correlate and refine temperature measurements across frames for them as they walk through. 




For tracking purpose, person tracking component maintains a cache, called as person cache, to store details of most recently detected individuals. Specifically, for these individuals, the latest bounding box location of body, head and face, along with their snapshots is maintained. A person remains in the cache until there are body, face or head detections available for the person. If there are no detections available for some configurable period of time e.g. 10 seconds, then the person is removed from the cache (since the person may have left and therefore there are no detections). In order to assign unique ID for an individual, each of the available detections i.e. body, face or head is checked one by one. 

Our tracking algorithm is described in Algorithm \ref{algo:tracking}. In the first loop, detected body is compared with all previously stored bodies in the cache to see if there is any match. For body match, we consider the bounding box location and image similarity. If the overlap in bounding box is high i.e. above a pre-configured threshold and the image similarity is also high (above a pre-configured threshold), then it is considered as a match. If a body match is found, then the ID of the body in cache is assigned to the detected body and details of the body i.e. bounding box and snapshot is updated in the cache. If there is no match found, then a new ID is assigned to the body and the detected body is added to the cache.

\begin{algorithm}
 \small
 \KwData{Body, Head, Face in frame}
 \KwResult{Body, Head, Face with tracking ID}
 \ForEach{detected body}{
 \eIf{body matches any body in person cache}{
   assign the ID in cache to the body\; 
 }{
   assign a new ID to the body\;
 }
}
 \ForEach{detected face}{
 \eIf{face matches a face in person cache}{
   assign the ID of the cache to the face\;
   \If{face in a body}{
    assign the ID of the face to the body\; 
   }   
 }{
  \eIf{face in a body}{
    assign the ID in body to the face\; 
   }{
    assign a new ID to the face\;
   }
  }
 }
  \ForEach{detected head}{
    assign ID of a face or body, or a new ID to the head;\
 }
\caption{Person tracking algorithm}
\label{algo:tracking}
\end{algorithm}

The second loop uses proprietary neural network based face recognition model \cite{neoface} to compare and match the face with all faces in the cache. If the match score is above a pre-configured threshold, then it is considered as a face match and the ID of the face in the cache is assigned to the detected face, and face details in cache is updated. Since accuracy of face recognition model is higher, we use this to overwrite the ID for body and assign the same ID as face, to the body. If there is no face match found, then ID of the body, if available, is assigned to the face, otherwise a new ID is assigned to the face and the detected face is added to the cache.

The third loop assigns an ID to the head detection. We do not use image similarity and bounding box location match for head because the head detection has a low detection rate compared to body and less accuracy for similarity match compared to face match. We assign the same ID as the face or body, if available, otherwise we assign a new ID to the head. If both face and body ID are available, then we give priority to face ID.

Thus by going through this procedure, where we check the available body, face or head detections one by one, we are able to track and assign unique ID to individuals, which enables \system\ to correlate and refine temperature measurement for an individual as the person walks through in a free-flow manner.

\subsection{Frame Fusion}
\lSec{frame-fusion}
As mentioned above, \system\ uses both, visual (RGB) and thermal sensors to achieve high accuracy and high throughput fever screening. The system detects persons on the visual frames and then finds the temperature of the person from the thermal frame. To do this, a mapping from the visual frame to the thermal frame is required, i.e. for a given point $(x_v, y_v)$ in the visual frame, what is the corresponding $(x_t, y_t)$ in the thermal frame. 
Now, $(x_v, y_v) == (x_t, y_t)$ if the sensors have the exact same viewpoint. However, even though these sensors are assembled in a single unit, they are placed side-by-side, with discernible distance between them. Such a setup introduces a difference in the sensor viewpoints, which implies that $(x_v, y_v) != (x_t, y_t)$. Instead, the points are related through a function - $(x_v, y_v) = f_{align}(x_t, y_t)$. The system needs to estimate $f_{align}$ so that a point from the visual frame can be mapped to the correct point in the thermal frame to read the associated temperature.



\subsubsection{Manual Offset Approach}

A straightforward way to get approximate alignment is to perform static scaling and translation. Let $I_{V}$ be visual sensor data, $I_{T}$ be thermal sensor data and $I_{AT}$ be thermal sensor data aligned to visual image. Then alignment function $f_{align}$ can be defined as follows (where $t_{x}$ is horizontal offset, $t_{y}$ is vertical offset, $S_{x}$ is horizontal scale factor and $S_{y}$ is vertical scale factor).

$f_{align}$ =
$\begin{bmatrix}
    1 & 0 \\
    0 & 1 \\
    t_{x} & t_{y}
\end{bmatrix}$
*
$\begin{bmatrix}
    S_{x} & 0 \\
    0 & S_{y}
\end{bmatrix}$

With above transformation function, it is possible to get temperature measurement in known area of visual ROI using data from $I_{AT} = f_{align}(I_{T})$.

However, this approach produces correct alignment only for a shallow depth plane, and approximate alignment for neighbouring pixels in that area of that plane. It does not align the entire frame. Pixels representing depth planes further or much closer from the correctly aligned image plane have larger misalignment.
Alignment errors can be about 100+ pixels (in both x and y direction) when the person is closer or further to camera. This leads to inaccurate temperatures being assigned to visual regions.

\subsubsection{Dynamic Frame Alignment}
\lSec{dynamic-frame-alignment}


\begin{figure}[b]
\centering
\includegraphics[width=0.9\linewidth]{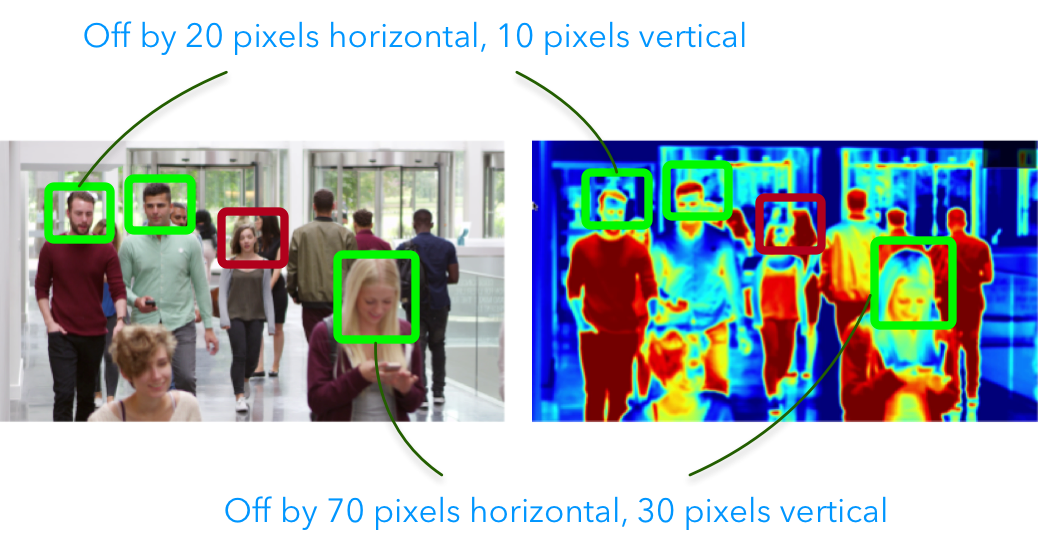}
\caption{Non-uniform misalignment}
\label{fig:non-uniform-misalignment}
\end{figure}

\begin{figure}[t]
\centering
\includegraphics[width=0.9\linewidth]{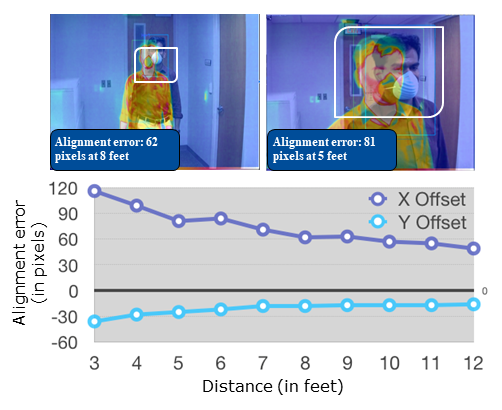}
\caption{Alignment error}
\label{fig:alignment-error}
\end{figure}

\begin{figure}[t]
\centering
\includegraphics[width=0.9\linewidth]{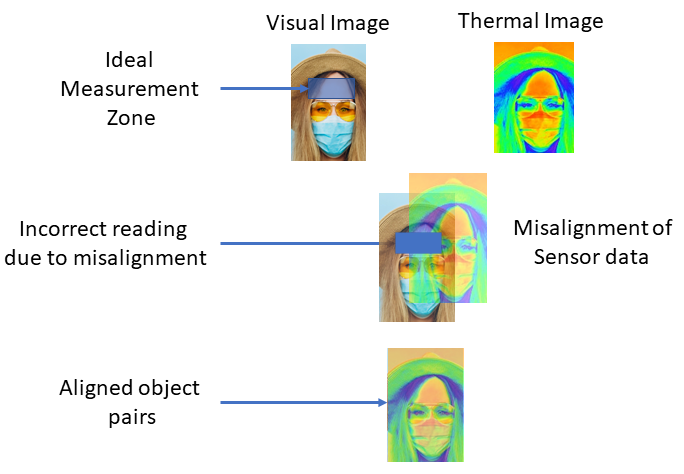}
\caption{Measurement inaccuracy due to Misalignment}
\label{fig:misaligned-images-and-dynamic-alignment}
\end{figure}

\begin{figure}[b]
\centering
\includegraphics[width=0.99\linewidth]{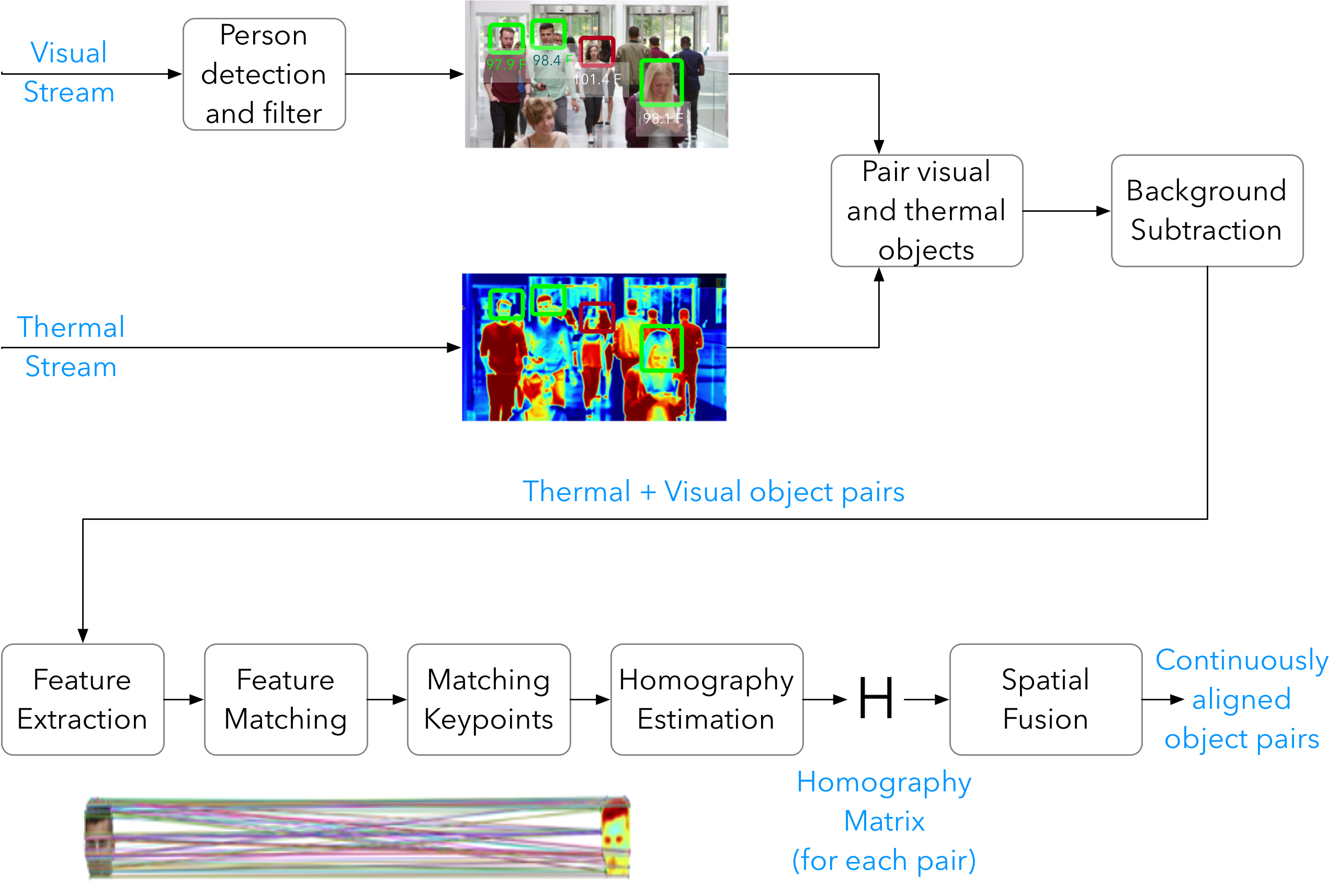}
\caption{Dynamic Frame Alignment}
\label{fig:dynamic-frame-alignment}
\end{figure}


Relative distortion between sensor images vary depending upon the depth plane. As the individual walks towards the camera, multiple temperature readings have to be taken to take into account head and face pose, occlusion with other individuals and objects. Since typical width of free flow traffic is about 6 to 8 feet wide and with horizontal field of view angle spans about 90 degrees or more (due to measurement zone constraints), it leads to non-uniform misalignment in horizontal plane as the individuals are found off camera's optical axis. Alignment issues is shown in Fig. \ref{fig:non-uniform-misalignment} and the alignment error at various distances along horizontal and vertical plane is shown in Fig. \ref{fig:alignment-error}. Corresponding measurement inaccuracy and required correction is depicted in Fig. \ref{fig:misaligned-images-and-dynamic-alignment}.

\system\ dynamically aligns visual and thermal image pairs for all people in field of view using the flow shown in Fig. \ref{fig:dynamic-frame-alignment}. Since thermal image has limited features to extract and limited correlation to visual image, features in boundary of person's head are matched in visual and thermal spectrum. To do this, first background is estimated in the images and subtracted and foreground mask is obtained. ORB features\cite{rublee2011orb} of foreground mask is obtained and matched using a brute force Hamming matcher to identify matching feature points. Using the matching feature points with high confidence score across both pairs, Homography matrix is obtained for each pair. Object pairs are rectified with respect to each other using Homography matrix in Spatial Fusion module.

A by-product of this alignment is distance measurement. The thermal and visual cameras are placed in a stereo arrangement with a small baseline. This allows the application of epipolar geometry and stereo vision principles to calculate depth-from-disparity \cite{szeliski2010computer}. Typically, the critical issue in assessing depth from cross-spectral stereo is finding corresponding points across the two spectra. We already solve this problem during the alignment process and using that, we obtain the depth and use it in subsequent module for temperature correction.

\subsection{Fever Screening}
Once visual and thermal frame fusion is completed, Fever-screening module then processes the fused frame to detect people with fever. Four key challenges that this module solves are:
\begin{enumerate}
\item Detecting and measuring temperature within the sensor's recommended measurement zone
\item Prioritizing temperature measurement region depending on available detections for an individual within a frame
\item Temperature correction based on distance from camera
\item Prioritized refinement of measured temperatures across frames, and alerting for those with fever
\end{enumerate}

\subsubsection{Detection and Measurement in Capture Zone.} \label{capture-zone}
\begin{figure}[b]
\includegraphics[width=\linewidth]{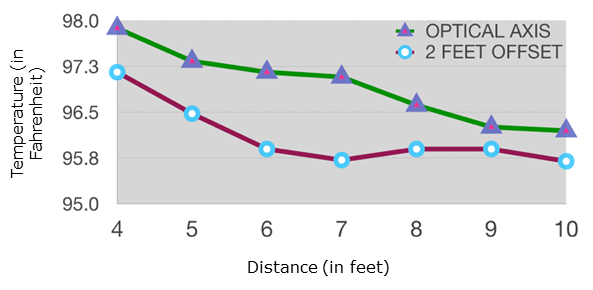}
\caption{Temperature variation with distance}
\label{fig:distance-vs-temperature}
\end{figure}
Measured temperature from the thermal sensor varies depending on the distance from the camera. This is because of the atmospheric composition between the person and the camera. Gases and particles present in the atmosphere absorb some of the emitted infrared radiation from the person. Farther the distance, more is the absorption and therefore less is the measured temperature. This variation of temperature with distance is shown in Fig. \ref{fig:distance-vs-temperature} for an individual along the optical axis and 2 feet off from the optical axis. Due to these atmospheric factors, thermal imaging sensors have a recommended measurement zone in terms of distance from the sensor. To allow measurement within this recommended zone, a capture zone is set up i.e. a zone in which temperature measurements for individuals starts as they enter and stops when they leave. Along with capture zone, a Region of Interest (ROI) within the camera view is also configured. This can be used to tightly control the region within which temperature is measured for individuals, thus keeping it clutter-free for the operator. Note that within the ROI, people may be too far or too near i.e. may be within or outside the capture zone.

\begin{algorithm}[t!]
 \small
 \KwData{Fused stream}
 \KwResult{Temperatures for detected people}
 discardIfFrameOutOfOrder()\;
 discardPersonsOutsideROI()\;
 \tcc{process detected persons}
 \ForEach{detected person within ROI}{
 \If{person within capture zone}{
 \eIf{person present in cache}{
   measureTemperatureAndUpdateInCache()\; 
 }{
 \uIf{person entered capture zone}{
   measureTemperatureAndAddInCache()\;
 }
 }
 }
 }
 renderFrameWithAnnotations()\;
 \tcc{calculate/revise temperature and send alert}
 sendTemperatureAlert()\;
 removeExpiredFromCache()\;

 \caption{Fever Screening algorithm}
  \label{algo:fever-screening-algorithm}
\end{algorithm}

Algorithm~\ref{algo:fever-screening-algorithm} shows the procedure followed in determining the temperature of individuals as they walk through in the capture zone. The first step is to discard out-of-order frames. Next, all detected persons who lie outside the configured ROI are removed. After removal of any person outside the ROI, the remaining individuals are processed one by one as they enter and leave the capture zone. 






Fever-screening module maintains a cache of recently seen individuals. Tracking ID of the individual is used as an identifier to determine if the person is new or previously seen. For a previously seen person, the new temperature reading is measured for the current frame and updated in cache corresponding to the tracking ID of the individual. For a new person, check is performed to see if the person has entered the capture zone. 
If it is determined that the person has entered the capture zone, then temperature for the person is measured and a new person with new tracking ID is added in cache. Note that the temperature for a person is added or updated in cache only if it is within the acceptable human temperature range, so as to avoid any spurious temperature readings. Also, note that we start temperature measurement for an individual when he/she enters the capture zone and continue to monitor and measure until the individual leaves the capture zone. 

After all individuals are processed, the frame is annotated with the temperatures for individuals and rendered, so that the operator can see the live view of the feed with temperatures of individuals annotated on the live stream. After frame rendering is done, all individuals in cache are processed one by one to (a) determine temperature for the individual across multiple readings and send alert if temperature is above a certain configured threshold and (b) remove any expired entries from cache i.e. remove any individuals in cache after a configurable period of time, after they have left.


\subsubsection{Prioritized temperature measurement.}
\begin{algorithm}[t]
 \small
 \KwData{Person detection within fused frame}
 \KwResult{Core body temperature for the person}
 $temperature \longleftarrow 0.0$\;
 $measured \longleftarrow false$\;
 \If{person has face detection} {
   \eIf{face has eye detection}{
     $temperature \longleftarrow getTemperatureAtEyeForeheadRegion()$\;
     $measured \longleftarrow true$\;
   }{
     $temperature \longleftarrow getTemperatureAtFaceRegion()$\;
     $measured \longleftarrow true$\;
   }
 }
 \If{NOT $measured$ AND person has head detection}{
     $temperature \longleftarrow getTemperatureAtHeadRegion()$\;
     $measured \longleftarrow true$\;
 }
 \If {$measured$} {
   $temperature \longleftarrow getCorrectedTemperature()$\;
 }
 \KwRet{$temperature$}
 \caption{Prioritized temperature measurement}
  \label{algo:per-frame-temperature-measurement}
\end{algorithm}

Algorithm~\ref{algo:per-frame-temperature-measurement} shows the procedure followed to prioritize and measure the temperature for a person within a frame. Highest priority is given to temperature measurement at eye and forehead region, followed by face and finally to head region. Within each of these regions, a configurable area is chosen and temperatures of pixels within that area is obtained from thermal sensor data. Among all the temperature values in the area, maximum among those is chosen as the raw measured temperature. We decided to choose the maximum value so as to avoid a false negative i.e. missing an individual who might have fever. 








\subsubsection{Temperature correction}

\begin{figure}[b]
\centering
\includegraphics[width=0.9\linewidth]{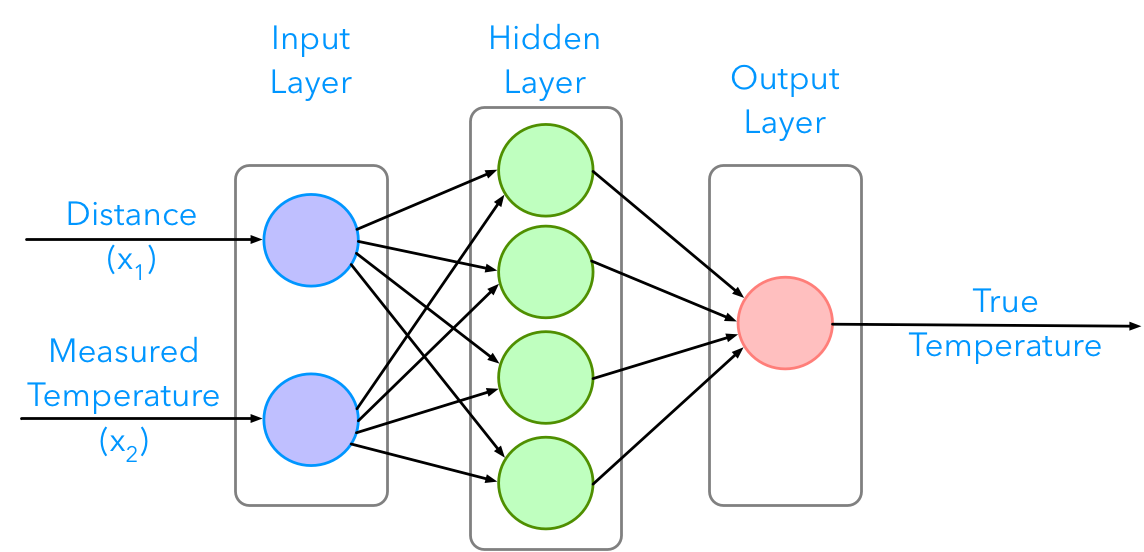}
\caption{Distance compensation model}
\label{fig:distance-compensation-model}
\end{figure}



As mentioned in section \ref{capture-zone}, the measured temperature of an individual varies depending on the distance from the camera and this variation is non-linear. To correct for this variation in measured temperature with distance, \system\ employs a neural network based distance compensation model, which can be used when a black body is present as part of the deployment. Black body provides a known reference temperature to ensure accurate temperature reading. This problem is framed as a regression problem and we use Multi-layer Perceptron (MLP) to solve it. The model is a feed-forward neural network, to which the input parameters are the distance and the corresponding measured temperature at that distance. The output from the model is the predicted true temperature of the individual. The model architecture is shown in Fig.  \ref{fig:distance-compensation-model}, with input layer, hidden layer and an output layer.

At the time of model training, as people walk through, the measured temperature at the black body is considered as the ground truth i.e. the true temperature of the person. Using this as the ground truth, the loss function is set as the mean squared error given by equation \ref{mean-squared-error}, where $Y_i$ is the true temperature and $\hat{Y_i}$ is the predicted temperature.

\begin{equation}\label{mean-squared-error}
    \mathscr{L} = \frac{1}{n}\sum_{i=1}^{i=n}{\Big(Y_i - \hat{Y_i}\Big)^2}
\end{equation}

We use Adam ~\cite{kingma2014adam} optimizer with following hyper-parameter settings: $\alpha$ of 0.001, $\beta_1$ of 0.9, $\beta_2$ of 0.999 and $\epsilon$ of $1e^{-07}$ for training. Rectified Linear Unit (ReLU) is used as activation function and the model is trained until 100 epochs with batch size of 1. The split between training and test data is 70:30. 
After the model is trained, the evaluation resulted in a mean squared error of 0.018 on test data, indicating a very high accuracy for the trained distance compensation model.

\subsubsection{Prioritized refinement and alerting}
As an individual walks through, multiple temperature readings of the person are recorded across frames. Based on these readings algorithm~\ref{algo:temperature-measurement-and-alerting} shows the procedure followed to prioritize and refine temperature measurement and send alert. To prioritize temperature measurement, highest priority is given to eyes and forehead, followed by face and head. Maximum temperature for the available highest priority region is recorded as the temperature for the person (to avoid missing a person with fever). To refine the measured temperature, as person is seen across frames, if temperature readings of higher priority region is measured, then the previous recorded temperature from lower priority region, is refined and updated with the new one. With respect to alerting, \system\ uses tracking ID to avoid sending repeated alerts for same individual. An alert is sent the first time measured temperature for a person is greater than the configured threshold, and after that an alert is sent only if higher priority region temperature is measured or if the delta increase in the temperature is above a configured threshold.

\begin{algorithm}[t]
 \small
 \KwData{Cache of recently seen individuals}
 \KwResult{Prioritized, refined temperature and alert}
 \ForEach{cache entry}{
   \If{minimum number of readings present} {
     $max\_head\_temperature \longleftarrow getMaxHeadTemperatureReading()$\;
     $max\_face\_temperature \longleftarrow getMaxFaceTemperatureReading()$\;
     $max\_eye\_forehead\_temperature \longleftarrow getMaxEyeForeheadTemperatureReading()$\;
     \eIf{higher priority region temperature is available}{
       updatePersonTemperatureWithHigherPriorityReading()\;
     }{
       updatePersonTemperatureReading()\;
     }
   }
   $delta\_temperature\_increase \longleftarrow getDeltaTemperatureIncrease()$\;
   \If{person temperature greater than configured threshold AND $delta\_temperature\_increase$ greater than configured delta change OR higher priority region temperature is available}{
     sendAlert()\;
   }
 }
 \caption{Prioritized refinement and alerting}
  \label{algo:temperature-measurement-and-alerting}
\end{algorithm}

\section{Auto-Calibration}
As the environmental conditions change, the thermal readings from the camera start to drift from actual temperature, thereby producing incorrect temperature readings, which ultimately results in incorrect temperature measurement for individuals. Camera vendors provide several parameters to calibrate and correct for this drift and maintain original temperatures. This correction however must be done manually, which is not practical to do in a real deployment where environmental conditions may change frequently. In order to do this automatically, \system\ employs a separate module to automatically detect the change and quickly adjust the camera parameters to maintain accurate temperature readings produced by the camera. 

\begin{figure}[b]
\centering
\includegraphics[width=0.9\linewidth]{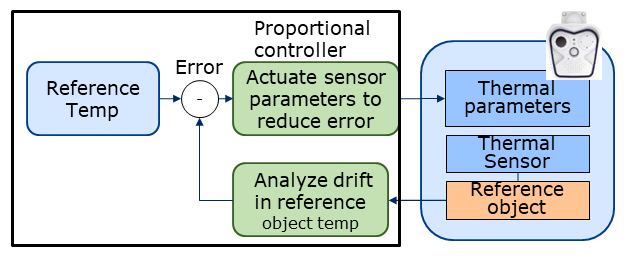}
\caption{Auto-calibration}
\label{fig:auto-calibration}
\end{figure}

\begin{figure}[t]
\centering
\includegraphics[width=0.9\linewidth]{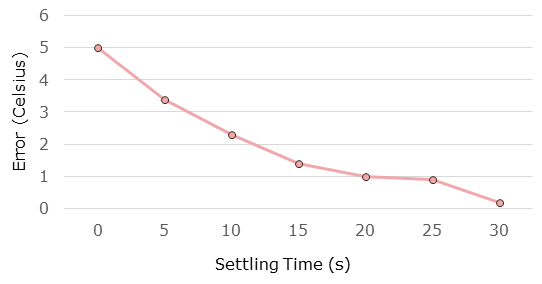}
\caption{Convergence of proportional controller}
\label{fig:convergence}
\end{figure}

In order to detect the change automatically, a black body is used as a reference object and set to a known reference temperature in the field of view of the camera. The temperature readings coming from the region of the black body is continuously monitored and any change in the temperature reading beyond an acceptable threshold is detected in real-time and immediate action is taken to adjust the camera parameters until the temperature of the black body is back to the reference temperature. Fig. \ref{fig:auto-calibration} shows a feedback control system used by \system\ for auto-calibration. The drift a.k.a. error in the temperature of the black body is measured in real-time and if the error is beyond an acceptable threshold, then the dynamic proportional controller corrects various parameters of the camera to iteratively reduce the error and bring back the measured temperature of the black body to the reference temperature. Using this auto-calibration technique, even if the environmental conditions change,  the camera always produces correct thermal readings, which ultimately results in correct temperature measurement for individuals. Fig. \ref{fig:convergence} shows the settling time for the proportional controller starting from an error of 5 degree Celsius i.e. the measured temperature of reference object is off by 5 degrees. A configurable time, called settling period is setup between each signal from the proportional controller, for the changed camera parameters to take effect. This is by default set to 5 seconds. For low errors, the convergence is achieved within a few signals, thereby maintaining accurate temperature measurement from the thermal sensor.

\section{Experimental Setup and Methodology}
In our system, we use an edge device with a quad-core processor, i7-7700, and 16GB memory. We use Ubuntu 16.04 as the OS and docker for software deployment. The hardware is able to process 8 frames per second (real-time performance) and the video frame resolution is 1280 by 960 while the resolution of thermal data is 336 by 252. We use a Mobotix M16 TR\cite{mobotix} for thermal and visual imaging in our experiments. Note that techniques used in \system\ are general and work with other cameras as well e.g. integrated thermal and visual cameras \cite{ici-fixed-thermal} or separate thermal \cite{seek-thermal} and visual \cite{intel-realsense} cameras placed next to each other.

\begin{table}[b]
\centering
\begin{tabular}{ |p{2.1cm}|p{1cm}|p{1cm}|p{1cm}|p{1cm}|} 
\hline
DATASET & \#people & \#mask & \#fever & length \\ 
\hline
LAB & 17 & 4 & 2 & 13m30s \\ 
\hline
PRODUCTION\_1 & 12 & 6 & 4 & 2m40s \\ 
\hline
PRODUCTION\_2 & 76 & 74 & 1 & 30m \\ 
\hline
\end{tabular}
\caption{\label{tab:dataset}Datasets used in our experiments}
\end{table}

We collect datasets containing both video and thermal data to measure the accuracy of our system. 
We split the task of assessing the accuracy of our system into two phases. First, we determine the accuracy of the thermal sensor by comparing the values it reports with  manual measurements using a traditional hand-held contact thermometer. Then, we assess the accuracy of \system, assuming that the thermal sensor is 100\% accurate. This two-step procedure has a big advantage: if we change the thermal sensor, then we only have to measure the accuracy of the new thermal sensor, and we can infer the accuracy of the system without further experiments.

For our experimental purpose, we created three datasets, called ``LAB", ``PRODUCTION\_1" and ``PRODUCTION\_2", as shown in Table \ref{tab:dataset}. ``LAB" is the one created in controlled lab environment while the other two are from production environment. In each dataset, we considered different operating conditions like standing at different distances and locations, walking at different speeds, wearing mask/glasses and having different temperatures, and we collected multiple video clips for these various conditions. The total number of people, those wearing masks, those with fever and total length of video clips for each dataset is shown in Table \ref{tab:dataset}.

Next, we discuss how we make ground-truth for these datasets. While for the ``LAB" dataset, we verify the thermal sensor's output with the data from a real thermometer and perform necessary tuning on the thermal sensor, in other datasets, we consider that the data from the thermal sensor is accurate. In other words, we use the raw data from thermal sensor as the ground-truth. To be more specific, for a person standing before the thermal sensor, we manually annotate the person's forehead and eye region and use the temperature of this area from the thermal sensor as the temperature of the person. 
Maximum temperature across frames is considered as ground-truth temperature for the person.


\section{Results}
\subsection{Dynamic Frame Alignment}
\begin{figure}[b]
\centering
\includegraphics[width=0.9\linewidth]{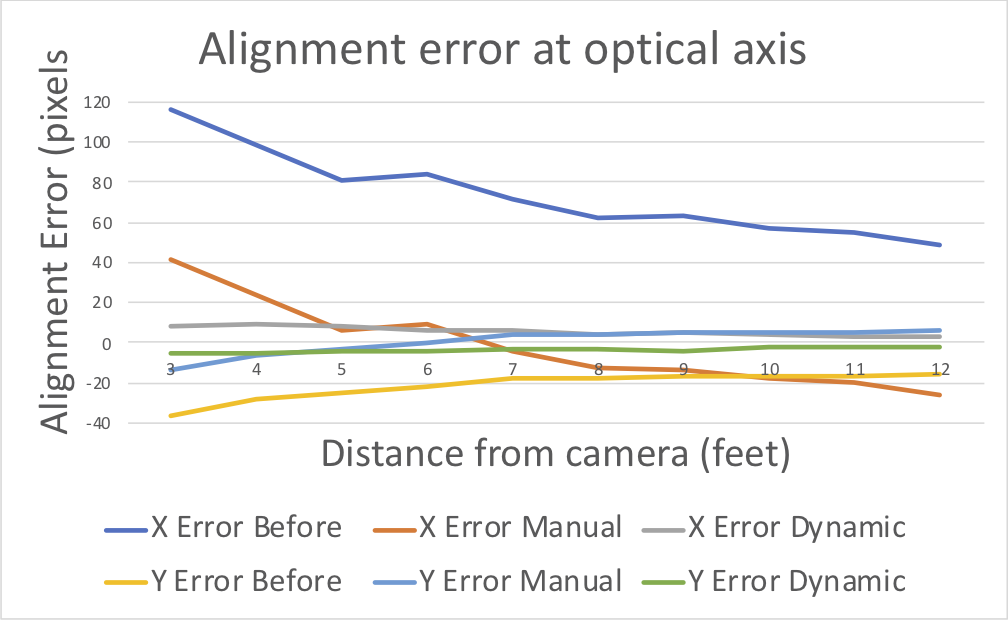}
\caption{Thermal and visual image alignment at optical axis}
\label{fig:thermal-and-visual-alignment-at-optical-axis}
\end{figure}

As explained in \rSec{frame-fusion}, due to displacement between thermal and visual sensor, the corresponding frames are not aligned, and hence can cause errors in mapping temperatures to visual objects. 
Here, we show how dynamic frame alignment achieves superior alignment as compared to the manual offset approach. As seen in Fig. \ref{fig:thermal-and-visual-alignment-at-optical-axis}, x co-ordinate alignment error varies from 120 pixels at 3 feet to 50 pixels at 12 feet (see ``X Error Before"). This is about 70 pixels of change as a person transits through the measurement area. Unfortunately, this is about the width of the bounding box for a face, and this causes erroneous readings of temperature for visual objects like face.

\begin{figure}[t]
\centering
\includegraphics[width=0.9\linewidth]{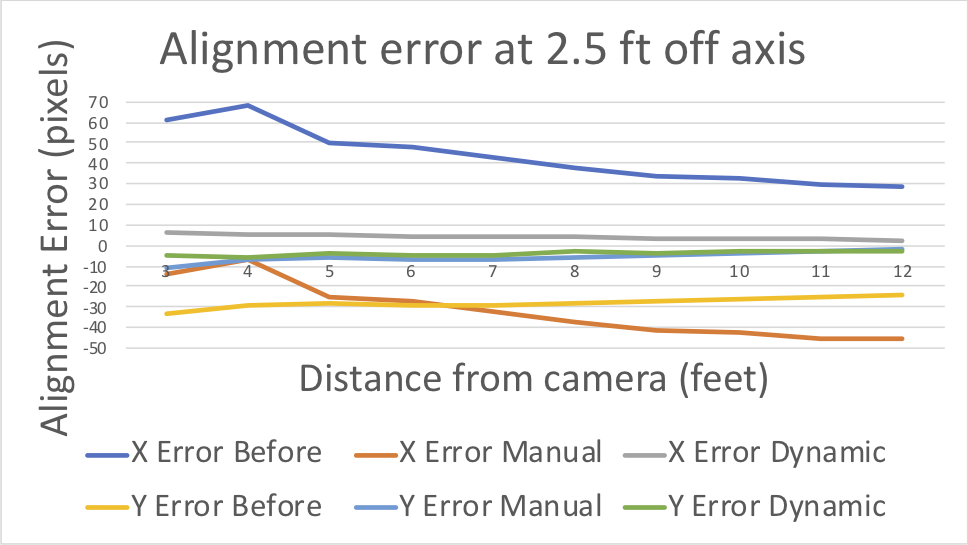}
\caption{Thermal and visual image alignment 2.5 ft off axis}
\label{fig:thermal-and-visual-alignment-at-2-5_feet-displacement}
\end{figure}

\subsubsection{Manual offset correction}
The manual x-offset and y-offset are set to 75 pixels and -25 pixels respectively in both Figs. \ref{fig:thermal-and-visual-alignment-at-optical-axis} and \ref{fig:thermal-and-visual-alignment-at-2-5_feet-displacement} based on manual overlay of the two images at a particular distance. These offsets are set once, independent of the position of the person in the field of view of the camera. This setting, however, does not provide good alignment at all distances from the camera as the person walks through (see ``X Error Manual" and ``Y Error Manual" in Figs. \ref{fig:thermal-and-visual-alignment-at-optical-axis} and \ref{fig:thermal-and-visual-alignment-at-2-5_feet-displacement}). For example, at 3 feet from the camera, and at the optical axis (Fig. \ref{fig:thermal-and-visual-alignment-at-optical-axis}), the x-coordinates of the visual and thermal region for a person are off by 120 pixels (``X Error Before"). Therefore, a manual x-coordinate correction of 75 pixels is not adequate to align the two frames. Similarly, a manual y-coordinate correction of -25 pixels is also not adequate for a person at 3 feet from the camera, where the error is -40 pixels (``Y Error Before").

\subsubsection{Dynamic correction}
By using dynamic frame alignment technique, both visual and thermal frames are tightly aligned at different distances, at and off optical axis (see ``X Error Dynamic" and ``Y Error Dynamic" in Figs. \ref{fig:thermal-and-visual-alignment-at-optical-axis} and \ref{fig:thermal-and-visual-alignment-at-2-5_feet-displacement}). For example, as shown in Fig. \ref{fig:thermal-and-visual-alignment-at-optical-axis}, after dynamic alignment, the x-coordinate alignment error (shown as ``X Error dynamic") for a person at 3 feet from the camera, and at the optical axis, is less than 5 pixels, which means that the visual and thermal frames are well aligned. The close alignment (low x and y alignment error) is observed for all usable distance ranges (3 feet to 12 feet) from the camera, and at the optical axis.  A similar near-perfect alignment is observed in Fig. \ref{fig:thermal-and-visual-alignment-at-2-5_feet-displacement}, where the dynamic alignment approach ensures that the x and y coordinate alignment error for all usable distances from the camera, and up to 2.5 feet off the optical axis,  is less than 5 pixels, indicating that the visual and thermal frames are well aligned.



\subsection{Temperatures at Various Regions}
\begin{figure}[b]
\centering
\includegraphics[width=0.9\linewidth]{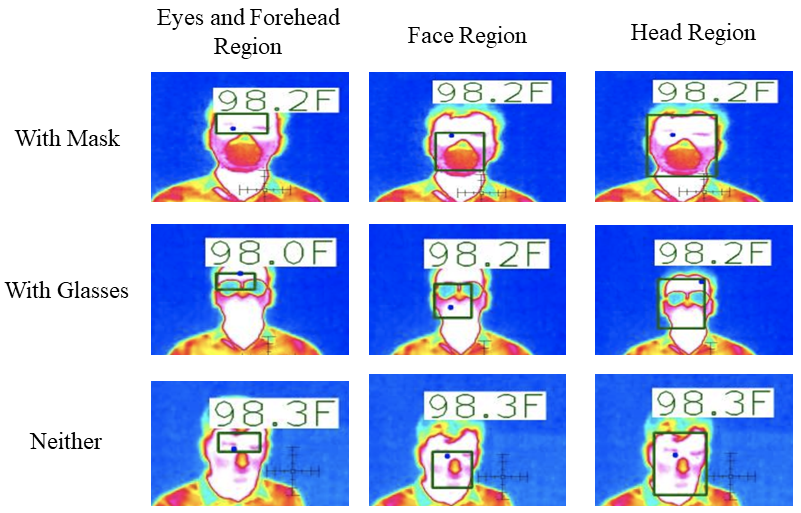}
\caption{Temperatures at various regions}
\label{fig:region-temperatures}
\end{figure}

Fig. \ref{fig:region-temperatures} shows our experimental results where we measure the temperature using detections at various regions i.e. eyes and forehead, face and head for a person wearing glasses, mask and neither. The rectangle shows the detections and dot inside the rectangle is the location of temperature measurement. We observe that when the person is not wearing glasses, across all detections, the final location of temperature measurement is the same. While, when the person is wearing glasses, the eye region is occluded and temperature is measured at different locations for different detections, but the variation in temperature is quite low. This shows that we do not need to ask people to pause and give a good shot, instead, we can use any of the available detections as they walk through. It typically takes $\sim$ 250 to 300 milliseconds for detection and temperature measurement for an individual.

\subsection{Fever Detection Accuracy}
\begin{table}[t]
\centering
\begin{tabular}{ |p{2.1cm}|p{1cm}|p{1cm}|p{1cm}|p{1cm}|} 
\hline
DATASET & TP & FP & TN & FN \\ 
\hline
LAB & 2 & 0 & 15 & 0 \\ 
\hline
PRODUCTION\_1 & 4 & 1 & 7 & 0 \\ 
\hline
PRODUCTION\_2 & 1 & 2 & 73 & 0 \\ 
\hline
OVERALL & 7 & 3 & 95 & 0 \\ 
\hline
\end{tabular}
\caption{\label{tab:results} \system\ Accuracy}
\end{table}

In Table \ref{tab:results}, we show the true positives (TP), false postives (FP), true negative (TN), and false negative (FN) from \system\ for each dataset. For a person with fever, if \system\ also detects fever for the person, it is considered as a TP, and if \system\ considers the person as normal, it is a FN case. For a person without fever, if \system\ detects fever for the person, it is a FP case, otherwise it is considered as a TN. For all three datasets with 105 people, \system\ achieves 100\% sensitivity and 96.9\% specificity. The standard deviation of temperature values is 1.41$^{\circ}$F. Among these datasets, we don't have any FN. In other words, we detects all people with fever, which is critical to guarantee the safety. However, our system has 3 false positives for PRODUCTION\_1 and PRODUCTION\_2 datasets due to occlusions and bad lighting conditions.

\section{Conclusion}
\lSec{conclusion}
We have presented a rapid, contact-less and hygienic fever screening system that runs at the edge and uses deep learning techniques for accurate temperature measurement. Our easy-to-use solution improves customer experience and works well even when individuals are using personal protective equipment like masks, spectacles and hats. Although not reported here, we have augmented the proposed system with face recognition. Locations with well-defined entry/exit points like building access, airport boarding gates, theme parks, stadiums and hospitals, all benefit from the high-speed and precision of the the proposed thermal screening and face recognition for access control and screening of employees and guests.
\bibliographystyle{IEEEtran}

\end{document}